\documentclass[11pt,a4paper]{article}
\usepackage[T1]{fontenc}
\usepackage{mathptmx,microtype}
\usepackage[margin=1in]{geometry}
\usepackage{amsmath,amssymb,graphicx,booktabs,array,longtable,placeins}
\usepackage{xcolor,titlesec,setspace,enumitem,fancyhdr}
\usepackage[round,authoryear]{natbib}
\definecolor{burgundy}{RGB}{128,0,32}
\usepackage[colorlinks=true,linkcolor=burgundy,citecolor=burgundy,urlcolor=burgundy,
pdftitle={Context-Dependent Affordance Reports in Vision-Language Models},pdfauthor={Murad Farzulla}]{hyperref}
\titleformat{\section}{\large\bfseries\color{burgundy}}{\thesection}{.6em}{}
\titleformat{\subsection}{\normalsize\bfseries\color{burgundy}}{\thesubsection}{.6em}{}
\graphicspath{{figures/}}
\newcommand{\code}[1]{\texttt{\small #1}}
\newcommand{\QAttempts}{3349}
\newcommand{\LAttempts}{3353}
\newcommand{\QParsed}{3213}
\newcommand{\QEmpty}{363}
\newcommand{\QEmptyPct}{11.3}
\newcommand{\QChefEmpty}{307}
\newcommand{\QMobilityEmpty}{55}
\newcommand{\QEmptyPairs}{2037}
\newcommand{\QPairs}{9244}
\newcommand{\QEmptyPairPct}{22.0}
\newcommand{\QWordOldPooled}{0.095}
\newcommand{\QWordNewPooled}{0.121}
\newcommand{\QCosOldPooled}{0.415}
\newcommand{\QCosNewPooled}{0.511}
\newcommand{\CosReplayError}{$4.17\times 10^{-7}$}
\newcommand{\QComplete}{115}
\newcommand{\LComplete}{437}
\newcommand{\EligibleImages}{4521}
\newcommand{\ChefLegacyLeverageRange}{0.9847--0.9849}
\newcommand{\ChefCompleteLeverageRange}{0.3607--0.3729}

\begin{document}
\begin{center}
{\small\textsc{Dissensus Working Paper Series}\quad DAI-2518}\\[1.2em]
{\LARGE\bfseries Context-Dependent Affordance Reports\\[.2em] in Vision-Language Models}\\[.6em]
{\large Separating Persona Effects from Measurement Artifacts}\\[1.3em]
{\large Murad Farzulla}\\[.4em]
Dissensus, London, UK\\
\href{mailto:murad@dissensus.ai}{murad@dissensus.ai}\quad
\href{https://orcid.org/0009-0002-7164-8704}{ORCID: 0009-0002-7164-8704}\\[.5em]
{\small 9 September 2026}
\end{center}

\begin{abstract}
Vision-language models produce different object and use descriptions under different persona prompts, but low overlap alone does not identify an affordance effect. We audit an earlier seven-prompt study and add matched-question controls. In the historical Qwen pilot, 363 of 3,213 parsed responses contain empty object lists. These affect 2,037 of 9,244 comparisons, with the implementation assigning zero lexical overlap to every affected pair. Conditioning on nonempty reports raises pooled word Jaccard from 0.095 to 0.121 and sentence cosine from 0.415 to 0.511. A previously named chef-specific Tucker factor loses its concentrated loading under missing-cell and complete-nonempty analyses. We withdraw the functional-manifold interpretation and the conversion of similarity scores into percentages of meaning. A new experiment uses 48 images absent from the original pilot, four personas, a shared three-object task, two wordings, and two requested seeds in each of two model configurations. Qwen3.5-9B's persona-minus-wording cosine-distance contrast is 0.0146 (95\% CI [-0.0003, 0.0295]; 37 complete images). Ollama llava:13b's persona-minus-wording cosine-distance contrast is -0.0266 (95\% CI [-0.0412, -0.0121]; 32 complete images).
 Persona-associated variation does not uniformly exceed wording or sampling variation. The study provides a reproducible analysis of context-conditioned reports while separating response availability, content similarity, and the limits of inference from linguistic outputs.

\end{abstract}

\vspace{1em}
\noindent\textbf{Revision disclosure.} This manuscript substantially revises
\href{https://arxiv.org/abs/2603.04419v2}{arXiv:2603.04419v2}. It corrects the treatment of empty responses, withdraws percentage-of-meaning and latent-manifold interpretations, and adds a prospectively specified matched-question experiment. The earlier public version remains a historical record. Appendix~\ref{app:corrections} identifies the replaced claims.

\section{Introduction}
An image can support several useful descriptions. A person inspecting a room for food preparation may report different objects and uses from someone assessing access through the same room. Vision-language models can produce such perspective-dependent reports. The empirical difficulty is deciding what an observed difference measures. A changed report may reflect a different question, a different selection of objects, another verbal formulation of the same use, sampling variation, or a failure to return content. These possibilities have different implications for an account of affordance computation.

The previous version of this study used seven persona prompts on the same images and reported low lexical overlap between responses. It interpreted one minus Jaccard similarity as the extent of context dependence, translated sentence cosine into a percentage of semantic content, and attached functional labels to a low-rank tensor decomposition. Reconstructing the analysis from raw outputs exposes a concrete failure: empty object lists were embedded and included in content comparisons even though the methods said they were excluded. An empty list generated no lexical tokens, so every comparison involving it received zero word overlap. Repeating this input through bootstrapping could stabilize a numerical factor without validating its functional interpretation.

This revision asks a narrower, operational question: \emph{when images and the reporting task are held fixed, do changes in stated persona produce more output variation than changes in wording or sampling seed?} It separates that prospective question from a retrospective audit of the original data. The audit is useful beyond this dataset because it makes visible three distinctions that an aggregate similarity score can obscure: whether an answer exists, what was selected for reporting, and how that report was worded.

The contribution has three parts. First, an executable response audit reconstructs the earlier results and quantifies their dependence on empty-response handling, output cardinality, image weighting, and metric choice. Second, a matched-question experiment crosses persona, wording, and requested seed on images absent from the historical pilot, with response-level provenance and image-level contrasts. Third, a claim-to-artifact record makes each retained conclusion traceable to raw outputs and a regenerator. The contribution is an empirical analysis of context-conditioned reports and their measurement.

\section{Affordances, reports, and related work}
\subsection{What the observation is}
An affordance is not identical to a sentence describing a possible use. In Chemero's relational account, affordances concern the relationship between an animal's abilities and environmental features, rather than a property located in either alone \citep{chemero2003}. The present data contain model-generated names and uses for objects in images. We call these \emph{affordance reports} to keep the linguistic observation distinct from physical feasibility, perception, and action. A reported use may be plausible, mistaken, role-stereotyped, or attached to an object absent from the image; lexical agreement alone cannot distinguish these cases.

A concrete illustration comes from historical image \code{000000000139.jpg}. The neutral report names a ``dining table'' and describes a flat surface for meals, work, or socializing; the security report names the same object and gives ``Barrier'' as its affordance. The reported object identity is shared, while the reported use changes. This example, selected for illustration, is not a scored judgment of feasibility or a representative estimate of the dataset. It shows why object selection and functional description merit separate measurements.

Context-dependent selection also has established antecedents. Computational models of task-dependent attention explicitly represent how a task can guide selection \citep{navalpakkam2005}. Controllable captioning can generate different descriptions by specifying image regions to describe \citep{cornia2019showcontroltell}. These results make a persona-dependent report unsurprising in principle. They do not establish the magnitude or reliability of the effect under this study's prompts, and they do not validate the old inference from output differences to the order of internal computation.

\subsection{Prompt sensitivity and physical grounding}
\citet{shah2025sensitivity} study VLM question sensitivity by adding modifiers to visual-question-answering prompts. Their outcome concerns answering performance, whereas ours concerns distances among free-form object/use reports. The shared lesson for experimental design is that prompt wording is an empirical source of variation and should have a measured baseline. Here, persona contrasts preserve the shared question within each wording version, and the two wording versions supply a within-persona comparator.

Physical grounding requires a different type of evidence. SayCan combines language-level task proposals with learned skill value functions to connect proposed actions to what a robot can execute \citep{ahn2022saycan}. Our static-image experiment has no robot, action policy, or execution outcome. Consequently, even a positive persona contrast would support a result about reported content, not improved embodied competence. Context-dependent reports also do not demonstrate that semantics precedes geometry internally. No activation, attention intervention, or causal probe of processing order is present here.

Affordance evaluation also covers several distinct targets. AffordanceLLM predicts interaction regions and evaluates grounding on AGD20K \citep{qian2024affordancellm}. A4Bench uses questions about constitutive and context-dependent affordances, including individual-specific cases \citep{wang2025a4bench}. \citet{rosenberg2026limits} compare human and VLM scene descriptions with a cosine-distance measure calibrated against within-human variability. These studies provide external criteria or human reference distributions that are absent here. Our within-model comparison instead asks whether role-conditioned variation exceeds wording and sampling variation; uncalibrated report similarity cannot supply the missing external criterion.

\section{Measurement framework}
\subsection{Availability precedes content comparison}
Let $Y_{ipwr}$ be a model response to image $i$, persona $p$, wording $w$, and requested random seed $r$. Let $R_{ipwr}$ indicate that a usable report is available. In the historical audit, a usable report must parse and contain a nonempty object list. In the new primary analysis it must also meet the required field and cardinality rules: exactly three distinct object names with a nonempty use for each.

The content estimand is conditional:
\begin{equation}
\mu_{ab}^{\rm content}=E\{d(Y_{ia},Y_{ib})\mid R_{ia}=R_{ib}=1\},
\label{eq:conditional}
\end{equation}
where $a,b$ index reporting conditions and $d$ is a prespecified output distance. The response probability $P(R_{ia}=1)$ is a separate outcome. If usability depends on persona or scene content, dropping failures changes the population of comparisons. Reporting a filtered mean therefore corrects an invalid content comparison but does not, by itself, identify an unconditional persona effect.

The historical word-level implementation assigned $J=0$ whenever either token set was empty. Its pooled mean consequently has the exact decomposition
\begin{equation}
\overline J_{\rm old}=(1-q)\overline J_{\rm both\ nonempty},
\label{eq:mixture}
\end{equation}
where $q$ is the fraction of comparisons involving an empty report. This equation describes that implementation's mixture; it is not a definition of similarity between an absent answer and an observed one. Sentence embeddings behave differently: an empty string receives a nonzero encoder vector, so empty-associated cosine values need not be zero.

An empty object list also has an ambiguous substantive meaning. It can be a schema-conforming response that no requested items were found, particularly when a prompt restricts the category. It can reflect inability to answer or other generation behavior. The records do not distinguish these mechanisms. We therefore use ``empty response'' rather than treating every empty list as a parsing failure, hallucination, or absence of real affordances.

\subsection{Output distances and their limits}
Raw-word Jaccard is the overlap of lowercased whitespace-token sets,
\begin{equation}
J(A,B)=\frac{|A\cap B|}{|A\cup B|},\qquad d_J=1-J.
\end{equation}
The resemblance measure is well defined for nonempty sets \citep{broder1997}. We also compare exact lowercased object-name sets. Neither measure resolves synonyms: ``couch'' and ``sofa'' are different names, and punctuation remains part of the raw-word metric to reproduce the historical analysis. This makes them useful descriptive checks, rather than validated semantic or grounding scores.

The primary prospective distance is $d_C=1-e(Y_a)^\top e(Y_b)$, with unit-normalized embeddings from the pinned \code{all-MiniLM-L6-v2} checkpoint. Sentence embeddings support cosine comparisons \citep{reimers2019sentence}; this particular encoder maps short text to 384 dimensions \citep{minilmcard}. We additionally embed names and uses separately. A cosine of 0.5 does not mean half the meaning is preserved. Nor are a unit of cosine distance and a unit of Jaccard distance interchangeable. The revision reports scores on their own scales and withdraws the earlier conversions into percentages of context-dependent meaning.

\subsection{Image-level aggregation and contrasts}
The pairs within an image reuse responses and are dependent. For each metric and pair type we first average within image, and then average the image means. An image with fewer usable comparisons consequently does not receive less weight merely because its responses generated fewer pairs. Pooled means remain in the audit only to reproduce the old report and demonstrate the zero-assignment mixture.

In the new experiment, define $D_i^P$ as mean distance between different personas at the same wording and seed; $D_i^W$ as mean distance between the two wordings at the same persona and seed; and $D_i^S$ as mean distance between the two seeds at the same persona and wording. The primary contrast and its sampling comparator are
\begin{equation}
\Delta_W=\frac{1}{N}\sum_i(D_i^P-D_i^W),\qquad
\Delta_S=\frac{1}{N}\sum_i(D_i^P-D_i^S).
\label{eq:contrast}
\end{equation}
Positive values mean that persona-associated report variation exceeds the chosen comparator. They are not percentages of explained variance. On a complete image there are 24 persona pairs, eight wording pairs, and eight sampling pairs; these yield one observation per contrast per image, not 40 independent observations.

For the prospective primary analysis, let $C_i=\prod_{p,w,r}R_{ipwr}$ indicate compliance of all 16 cells. Equation~\ref{eq:contrast} averages only images with $C_i=1$ and therefore estimates $E(D_i^P-D_i^W\mid C_i=1)$, with an analogous sampling contrast. This is stricter conditioning than the available-pair content quantity in Equation~\ref{eq:conditional}; its advantage is matched support across comparison types, and its cost is exclusion of every image with any unusable cell.

We report two-sided 95\% $t$ intervals on image means and paired image differences. These intervals describe uncertainty across images for the specified personas, wordings, and requested seeds. Two wordings are insufficient to generalize to all paraphrases, and two seeds do not characterize an entire sampling distribution. We use no arbitrary $J=0.5$ hypothesis test and no pooled-pair significance test. Metric-specific intervals are descriptive; the primary interpretation is attached to cosine distance and $\Delta_W$ separately for each model.

\section{Historical data audit}
\subsection{Sources, prompts, and provenance}
The archived Qwen pilot contains \QAttempts{} recorded attempts on 479 COCO-2017 validation images \citep{lin2014coco}. Four of the $479\times7$ image/prompt cells have no record in the archive; these absences are separate from recorded responses that fail parsing. Its collection script names Qwen3-VL-30B-A3B, but the individual records do not pin model weights or backend version. We retain that label as the reported collection identity. Seven prompts specify an adult observer, chef, security professional, child, wheelchair user, an urgent survival task, and leisure. They change both perspective and requested content. Most ask for three objects; the urgent prompt is singular, and the leisure prompt leaves cardinality unspecified. These differences prevent a pure persona interpretation of the historical cross-prompt contrast.

The archived comparison contains \LAttempts{} attempts from the Ollama tag \code{llava:13b}. The earlier paper called this LLaVA-1.5-13B, but no historical model digest is stored. The tag's present identity cannot establish its identity at the time of collection. Historical results are therefore labeled by the recorded tag, without assigning a release or explaining between-model differences by architecture. The new experiment records the current digest explicitly.

The parser removes Markdown JSON fences and attempts a direct JSON decode; it performs no repair or substring recovery. Non-dictionary objects, non-list object fields, and non-string text fields are schema errors. Duplicate image/persona keys are rejected rather than silently overwritten. Text concatenates object name, affordance, and reasoning fields in their original order, matching the earlier implementation. The response ledger retains every original attempt and its classification.

\subsection{Missing answers account for part of the apparent divergence}
\begin{table}[htbp]\centering\small
\caption{Historical response availability.}
\label{tab:availability}
\begin{tabular}{lrrrr}
\toprule
Collection & Attempts & Nonempty & Empty & Parse/schema errors \\
\midrule
Qwen pilot & 3349 & 2850 & 363 & 136 \\
Ollama tag & 3353 & 3307 & 0 & 46 \\
\bottomrule\end{tabular}

\end{table}

Of \QParsed{} parsed Qwen reports, \QEmpty{} contain an empty object list (\QEmptyPct{}\%). Of those empty lists, \QChefEmpty{} occur under the chef prompt, \QMobilityEmpty{} under mobility, and one under urgency. They affect \QEmptyPairs{} of \QPairs{} available comparisons (\QEmptyPairPct{}\%). All zero word-Jaccard pairs in this historical dataset involve empty text. At the object-name level, many additional zero overlaps occur between nonempty lists; the empty-response correction does not explain away all object-selection differences.

\begin{table}[htbp]\centering\small
\caption{Historical and corrected similarity estimates.}
\label{tab:legacy}
\begin{tabular}{lrrll}
\toprule
Collection / support & Images & Pairs & Word Jaccard [95\% CI] & Cosine [95\% CI] \\
\midrule
Qwen / All parsed & 479 & 9244 & 0.094 [0.092, 0.097] & 0.414 [0.406, 0.421] \\
Qwen / Nonempty & 479 & 7207 & 0.121 [0.119, 0.122] & 0.513 [0.506, 0.519] \\
Qwen / Complete & 115 & 2415 & 0.125 [0.123, 0.127] & 0.504 [0.491, 0.517] \\
Ollama / All parsed & 479 & 9787 & 0.160 [0.158, 0.163] & 0.586 [0.577, 0.595] \\
Ollama / Nonempty & 479 & 9787 & 0.160 [0.158, 0.163] & 0.586 [0.577, 0.595] \\
Ollama / Complete & 437 & 9177 & 0.160 [0.158, 0.163] & 0.587 [0.577, 0.596] \\
\bottomrule\end{tabular}
\par\smallskip\begin{minipage}{.98\linewidth}\footnotesize Means and intervals weight images equally. Complete denotes images with all seven nonempty reports. All parsed reproduces the empty-response convention; it is not the preferred content estimand. Object-name and cached lemmatized metrics are available in the machine-readable analysis.\end{minipage}
\end{table}

The reconstructed pooled word-Jaccard mean is \QWordOldPooled{}, and the pooled mean conditional on two nonempty answers is \QWordNewPooled{}. Equation~\ref{eq:mixture} exactly accounts for their difference. The corresponding sentence-cosine means are \QCosOldPooled{} and \QCosNewPooled{}. These cosine values are regenerated from raw text; the largest absolute discrepancy from the archived pair scores is \CosReplayError{}. Table~\ref{tab:legacy} uses image-weighted means and intervals consistently, so its values differ slightly from the pooled historical headline.

The correction changes the estimated between-model gap as well. The historical Ollama comparison has no empty object lists among its parsed reports. Comparing its lexical overlap to Qwen's unfiltered overlap combines content behavior with a strongly asymmetric empty-response profile. The remaining gap after conditioning is still descriptive: prompts, decoding backends, quantization, output wording, and historical model identity do not support an architectural attribution.

\begin{figure}[tbp]
\centering\includegraphics[width=\linewidth]{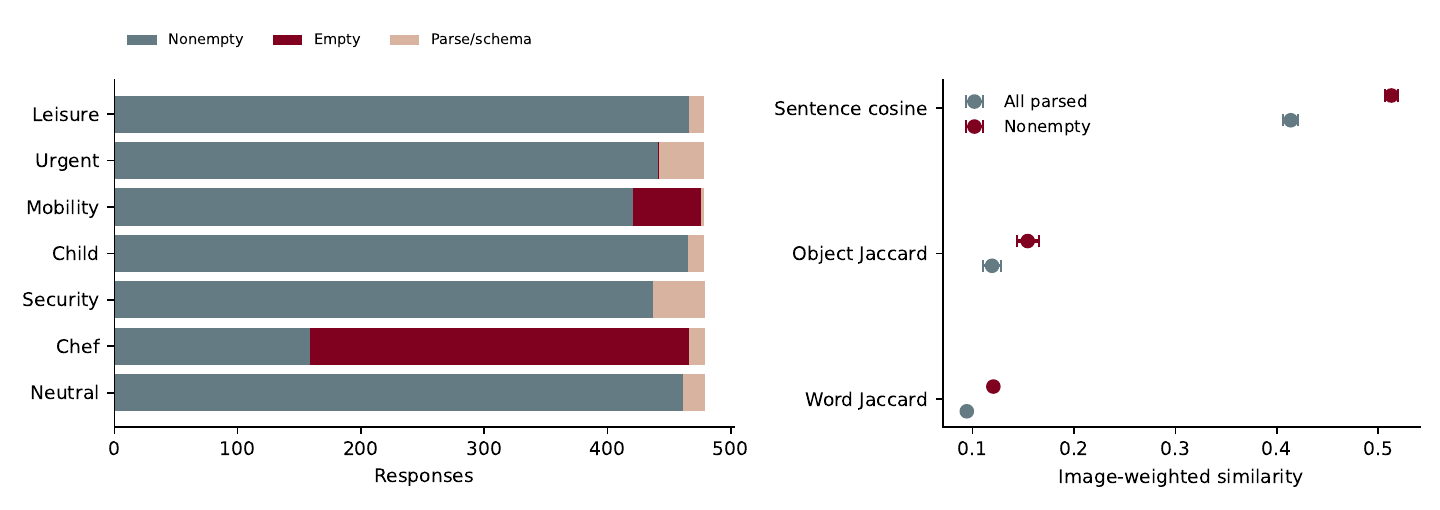}
\caption{Response availability and metric sensitivity in the historical Qwen pilot. Left: outcomes by prompt. Right: image-weighted means and 95\% intervals before and after restricting comparisons to nonempty responses. The metrics have different meanings and are not percentages of a common semantic quantity.}
\label{fig:audit}
\end{figure}

\subsection{Cardinality and complete-case sensitivities}
Exactly \QComplete{} images contain seven nonempty Qwen reports. Their complete set of 21 comparisons per image provides a balanced-persona sensitivity, but it selects images that support every prompt and cannot substitute for the full 479-image population. Table~\ref{tab:legacy} reports this subset separately. The same rule retains \LComplete{} images in the historical Ollama data.

For a cardinality sensitivity, we hold the pair universe fixed at reports with at least three listed objects in both responses. We then compare prefixes of depth one, two, and three. This removes a mechanical difference in the number of listed objects within each comparison, while preserving the same support across depths. It does not repair the original prompt confound or identify causal effects of asking for fewer objects: prefixes of an already generated list are not fresh generations under a shorter-list instruction. The generated tables also retain a larger available-support version at each depth for audit purposes.
\begin{table}[htbp]\centering\small
\caption{Qwen prefix sensitivity on fixed pair support.}
\label{tab:cardinality}
\begin{tabular}{rrrll}
\toprule
Depth & Images & Pairs & Word Jaccard [95\% CI] & Object Jaccard [95\% CI] \\
\midrule
1 & 479 & 5934 & 0.122 [0.120, 0.124] & 0.291 [0.266, 0.316] \\
2 & 479 & 5934 & 0.122 [0.121, 0.124] & 0.197 [0.181, 0.212] \\
3 & 479 & 5934 & 0.121 [0.120, 0.123] & 0.164 [0.152, 0.177] \\
\bottomrule\end{tabular}
\par\smallskip\begin{minipage}{.98\linewidth}\footnotesize The same pair universe has at least three listed objects in both reports. These are prefixes of existing answers, not new shorter-list generations.\end{minipage}
\end{table}

On this fixed support, raw-word overlap changes little across prefix depths, whereas exact-name overlap decreases as the lists include additional objects. The two metrics therefore respond differently to the same cardinality operation. A general claim that shorter lists must inflate divergence is not supported by this sensitivity; the empirical direction depends on what is compared and how the sets expand.

\subsection{The latent-factor interpretation fails a missing-response check}
The previous Tucker analysis used a $360\times7\times384$ tensor built from images with seven parsed responses, including empty lists. Of its 2,520 cells, 280 contain blank text, including 238 chef cells. Thus roughly two thirds of the chef slice represent the same empty-string embedding. Tucker decomposition \citep{tucker1966} can recover that regularity reliably without recovering a domain-specific representation of cooking.

We reproduce the original rank $(10,3,10)$ fit using TensorLy 0.9.0, HOOI, random initialization, seeds 42--44, a 100-iteration cap, and tolerance $10^{-4}$. We compare two sensitivity fits: the 115 images with seven nonempty reports, and all 360 images with blank cells masked as missing. The masked fit uses TensorLy's masked SVD initialization with missing input cells initially zeroed, and a 300-iteration cap; its completion assumes low-rank structure for the unobserved cells and is not recovered ground truth. An unstable random-initialized masked fit is retained in the runtime record and excluded from substantive interpretation.

\begin{table}[htbp]\centering\small
\caption{Tucker factor sensitivity to empty-response handling.}
\label{tab:tucker}
\begin{tabular}{lrrrr}
\toprule
Input & Images & Energy capture & Chef on old axis 2 & Chef leverage \\
\midrule
Blank embeddings & 360 & 0.466 & 0.954 & 0.985 \\
Masked blanks & 360 & 0.452 & 0.324 & 0.243 \\
Complete nonempty & 115 & 0.481 & 0.474 & 0.373 \\
\bottomrule\end{tabular}
\par\smallskip\begin{minipage}{.98\linewidth}\footnotesize Seed 42 shown; the archive contains seeds 42--44. Masked SVD initialization is deterministic here, so repeated seed arguments are not independent starts. Other fits use random initialization. Axis matching and completion assumptions are described in the text. Neither energy capture nor row leverage is a semantic percentage.\end{minipage}
\end{table}

The published-style fit reproduces the near-exclusive chef loading. That concentration falls in both nonempty sensitivities. Loadings in Table~\ref{tab:tucker} are matched by maximum absolute congruence to the three previously published columns, with sign alignment, so displayed axes remain reference-dependent. As an additional rotation-invariant check, we report chef row leverage $\|U_{\mathrm{chef},:}\|_2^2$, where $U$ is the three-column persona factor matrix. This is the diagonal of the fitted subspace projector, not evidence of a functional manifold.

Across the three random starts, chef leverage ranges from \ChefLegacyLeverageRange{} in the historical input and \ChefCompleteLeverageRange{} in the complete-nonempty input. The reduced concentration therefore does not depend on selecting seed 42. Masked initialization is deterministic in this implementation; repeating its seed argument does not provide independent initialization evidence.

The masked analysis retains the same image support as the original tensor but imposes a completion model; the complete-case analysis avoids imputation but changes the image population. Their agreement on reduced chef concentration undermines the earlier interpretation without identifying a unique causal decomposition of the artifact. We withdraw the ``Culinary Manifold'' and ``Access Axis'' claims. We also withdraw orthogonality as evidence: HOOI factor columns are orthonormal by construction. The fit quantity $1-\|T-\widehat T\|_F^2/\|T\|_F^2$ is uncentered reconstruction-energy capture, not centered variance explained.

\section{Matched-question experiment}
\subsection{Design frozen before collection}
The new experiment samples 48 COCO validation images without replacement from the \EligibleImages{} local images absent from the historical Qwen pilot. Image selection uses seed 20260908. The sample size was fixed to fit the local collection budget, without an a priori power calculation. The exact image list and SHA-256 hashes, prompts, and randomized schedule were committed before the first scheduled request; the protocol commit is \code{076395b}. This is prospective local specification, not external preregistration. Technical smoke tests used a historical image excluded from the new sample.

Each image is queried under four personas: adult observer, professional chef, security professional, and wheelchair user. Within each of two wording versions, the task is identical across personas: name exactly three visible objects, rank their relevance to the stated perspective, and give one plausible use for each. Every condition permits an ordinary use when a role-specific use is unavailable and instructs the model not to invent unseen objects to fit the role. This shared-category design removes the explicit cooking/security/mobility category restrictions of the historical prompts. It does not make persona meanings interchangeable or establish the correctness of a report.

Each persona/wording cell is run with requested seeds 11 and 29 at temperature 0.7. This yields $48\times4\times2\times2=768$ scheduled requests per model, each with a fresh image-and-text request and a 256-token output limit. The two wording variants paraphrase both the role sentence and the shared question. Their pairwise comparison is a measured nuisance contrast over these two versions, not a claim that they are semantically equivalent for every model.

\subsection{Models, generation, and execution record}
We use the locally installed Qwen3.5-9B Q8\_0 model through LM Studio \citep{qwen35card}, and the current Ollama \code{llava:13b} Q4\_0 model \citep{ollamallava}. Both have a 4,096-token context setting. Qwen thinking is disabled; the raw API responses retain the returned reasoning-token counts so this is checkable. Requested seeds are recorded, but backend seeds are not assumed to guarantee bitwise replay. Raw responses, generation parameters, finish reasons, timestamps, model metadata, and model-file or manifest hashes accompany the data.

The original Qwen3-VL-30B weights were not available for the new experiment, so this is a controlled extension with a newer model, not a matched re-inference of the historical Qwen collection. The current Ollama digest identifies the new run but does not retrospectively identify the old one. Model, backend, and quantization are bundled within each experimental arm; any observed difference between arms concerns these configurations as deployed.

An initial concurrent execution caused GPU contention. Collection was serialized between model backends, and Qwen concurrency was reduced from two requests to one. Previously persisted responses were retained, and only scheduled cells without a stored response were resumed. The exact operational restart and completion records are in the archive. Randomized order limits systematic time ordering of persona cells, but hardware scheduling remains part of the execution environment.

\subsection{Response validation and analysis support}
Primary content analysis requires all 16 scheduled reports for an image to parse and contain three distinct, nonempty names, each with a nonempty use. This ensures that persona, wording, and sampling contrasts use identical images and cells. The validator checks the required content but ignores additional fields, such as an explicit rank; these fields do not enter the text representation. The counts of transport failures, parse/schema errors, empty answers, incomplete fields, repeated names, incorrect cardinality, and token-limit stops are reported separately. We neither replace failed responses nor repair their JSON in the analysis.

An available-pair sensitivity retains all nonempty responses with usable names and uses, including incorrect list lengths, and averages the available pairs within image. This can include different cell supports across the three pair types and is therefore secondary. The primary complete-case contrast can also be selected if compliance depends on image or persona. Its population is the set of compliant image blocks, not necessarily all sampled images.

The primary embedding concatenates names and uses in listed order and lowercases the resulting text; secondary embeddings use names alone and uses alone. Uses-only comparisons may still involve different selected objects, so they do not isolate a persona effect on the use assigned to a fixed object. Exact-name and raw-word Jaccard distances supply additional lexical checks. Embeddings preserve listed order, whereas the set-based checks discard it; a cosine difference can therefore include reordering as well as changed content and is not a calibrated ranking score. To examine retained image signal, we compare same-image/different-persona distances with different-image/same-persona distances, holding wording and requested seed fixed. Each complete image is compared with all other complete images. Their dependence makes this a descriptive diagnostic; it is not a hallucination accuracy score or an independent-image-pair test.

Before computing outcome summaries, a supplementary specification added three checks: re-embedding full reports with the pinned \code{all-mpnet-base-v2} encoder on the identical complete-image support; 10,000 percentile-bootstrap resamples of images, with seed 20260910, and leave-one-image-out means; and a paired comparison of model configurations on their shared complete images. These checks assess encoder choice, interval construction, and support. They do not replace the MiniLM primary contrast when a result differs. The supplementary specification is recorded in local commit \code{5d30151}.

\subsection{Results}
\begin{table}[htbp]\centering\small
\caption{Matched-question experiment: response compliance.}
\label{tab:control-counts}
\begin{tabular}{lrrrr}
\toprule
Model & Attempts & Compliant & Other outcomes & Complete images \\
\midrule
Qwen3.5-9B & 768 & 757 & 11 & 37 \\
Ollama llava:13b & 768 & 744 & 24 & 32 \\
\bottomrule\end{tabular}
\par\smallskip\begin{minipage}{.98\linewidth}\footnotesize A complete image has 16 compliant responses. Other outcomes remain in the response ledger and exclude an image from the primary analysis. The available-pair sensitivity retains usable cardinality or duplicate-name violations, as specified in the text.\end{minipage}
\end{table}

Qwen3.5-9B retains 37 complete images. Mean cosine distance is 0.315 [0.290, 0.340] across personas, 0.300 [0.277, 0.324] across wording versions, and 0.295 [0.273, 0.316] across requested seeds. The paired persona-minus-wording contrast is 0.0146 [-0.0003, 0.0295], and persona-minus-sampling is 0.0205 [0.0095, 0.0314]. Brackets give image-level 95\% confidence intervals.

Other outcomes for Qwen3.5-9B: empty (8), parse error (3).

In the image comparison, Qwen3.5-9B's mean different-image/same-persona cosine distance is 0.861, compared with 0.315 for same-image/different-persona pairs. This is a descriptive check of image-associated variation, not evidence that the named objects or uses are correct.

Ollama llava:13b retains 32 complete images. Mean cosine distance is 0.284 [0.255, 0.313] across personas, 0.311 [0.284, 0.337] across wording versions, and 0.341 [0.317, 0.365] across requested seeds. The paired persona-minus-wording contrast is -0.0266 [-0.0412, -0.0121], and persona-minus-sampling is -0.0569 [-0.0754, -0.0384]. Brackets give image-level 95\% confidence intervals.

Other outcomes for Ollama llava:13b: cardinality violation (14), duplicate names (3), parse error (7).

In the image comparison, Ollama llava:13b's mean different-image/same-persona cosine distance is 0.859, compared with 0.284 for same-image/different-persona pairs. This is a descriptive check of image-associated variation, not evidence that the named objects or uses are correct.

\begin{table}[htbp]\centering\small
\caption{Persona-associated variation relative to wording and sampling baselines.}
\label{tab:contrasts}
\begin{tabular}{llll}
\toprule
Model & Distance & Persona minus wording & Persona minus sampling \\
\midrule
Qwen3.5-9B & Full cosine & 0.0146 [-0.0003, 0.0295] & 0.0205 [0.0095, 0.0314] \\
 & Names cosine & 0.0152 [-0.0053, 0.0358] & 0.0186 [0.0044, 0.0329] \\
 & Uses cosine & 0.0090 [-0.0094, 0.0275] & 0.0145 [0.0009, 0.0280] \\
 & Word Jaccard & 0.0049 [-0.0063, 0.0160] & 0.0026 [-0.0086, 0.0138] \\
 & Object Jaccard & 0.0025 [-0.0207, 0.0257] & -0.0134 [-0.0368, 0.0100] \\
Ollama llava:13b & Full cosine & -0.0266 [-0.0412, -0.0121] & -0.0569 [-0.0754, -0.0384] \\
 & Names cosine & -0.0215 [-0.0390, -0.0040] & -0.0510 [-0.0728, -0.0293] \\
 & Uses cosine & -0.0363 [-0.0561, -0.0166] & -0.0660 [-0.0883, -0.0437] \\
 & Word Jaccard & -0.0506 [-0.0708, -0.0303] & -0.0717 [-0.0952, -0.0482] \\
 & Object Jaccard & -0.0586 [-0.0965, -0.0206] & -0.1461 [-0.1912, -0.1010] \\
\bottomrule\end{tabular}
\par\smallskip\begin{minipage}{.98\linewidth}\footnotesize Contrasts use identical complete image blocks within each model. The full-text cosine contrast against wording is primary. Secondary metric intervals are descriptive, not multiplicity-adjusted tests. Positive values mean larger between-persona distance on that metric.\end{minipage}
\end{table}

For Qwen3.5-9B, the available-pair sensitivity uses 48 images and gives persona-minus-wording 0.0117 [-0.0009, 0.0242]. For Ollama llava:13b, the available-pair sensitivity uses 48 images and gives persona-minus-wording -0.0153 [-0.0299, -0.0008]. The available-pair analysis may change cell support across comparison types; it is not the primary matched-support contrast.

\begin{figure}[tbp]\centering\includegraphics[width=.92\linewidth]{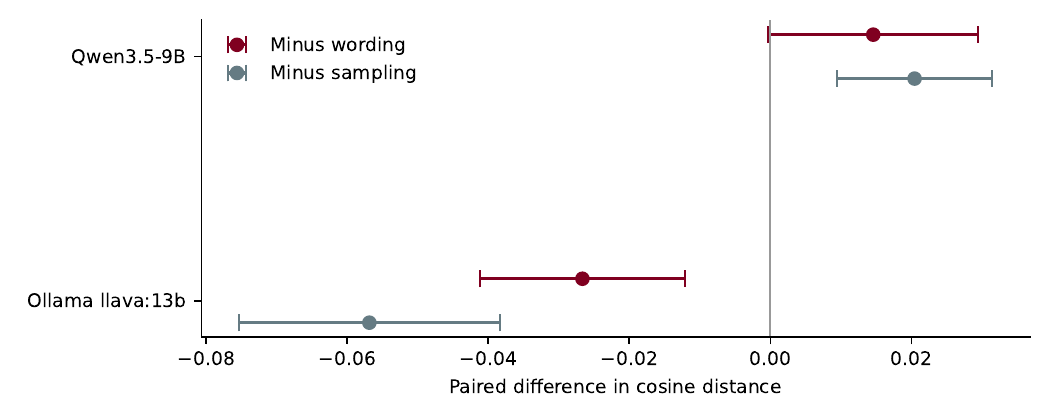}\caption{Image-level mean persona-minus-baseline cosine distances and 95\% intervals. Zero denotes equality with the particular wording or sampling comparator, not absence of all context effects.}\label{fig:controls}\end{figure}

\subsection{Sensitivity to encoder, resampling, and shared support}
\begin{table}[htbp]\centering\small
\caption{Sensitivity of the persona-minus-wording contrast.}
\label{tab:sensitivity}
\begin{tabular}{llrl}
\toprule
Model & Encoder / interval & Images & Contrast [95\% CI] \\
\midrule
Qwen3.5-9B & MPNet / t interval & 37 & 0.0156 [0.0016, 0.0296] \\
 & MiniLM / bootstrap & 37 & 0.0146 [0.0010, 0.0290] \\
Ollama llava:13b & MPNet / t interval & 32 & -0.0243 [-0.0386, -0.0100] \\
 & MiniLM / bootstrap & 32 & -0.0266 [-0.0407, -0.0131] \\
\bottomrule\end{tabular}
\par\smallskip\begin{minipage}{.98\linewidth}\footnotesize Both encoders use the same complete images within a model. Bootstrap intervals resample images 10,000 times. These are supplementary analyses, not additional independent experiments.\end{minipage}
\end{table}

For Qwen3.5-9B, leave-one-image-out MiniLM persona-minus-wording means range from 0.0106 to 0.0167. The MPNet persona-minus-sampling contrast is 0.0209 [0.0086, 0.0332].

For Ollama llava:13b, leave-one-image-out MiniLM persona-minus-wording means range from -0.0289 to -0.0238. The MPNet persona-minus-sampling contrast is -0.0464 [-0.0645, -0.0283].

On the 25 images complete in both configurations, Qwen minus Ollama gives 0.0411 [0.0119, 0.0702] for the persona-minus-wording contrast. This paired contrast compares the configurations as deployed.

On the 25 images complete in both configurations, Qwen minus Ollama gives 0.0787 [0.0494, 0.1080] for the persona-minus-sampling contrast. This paired contrast compares the configurations as deployed.

For Qwen3.5-9B, the persona-minus-wording point estimate is similar under both encoders, but the interval boundary is sensitive to the analysis choice. The primary MiniLM $t$ interval includes zero; the MPNet $t$ interval and the MiniLM percentile-bootstrap interval have lower bounds slightly above zero. The estimated excess is modest on these distance scales, and its distinction from zero is borderline. No equivalence margin was specified, so an interval including zero also does not establish that the two sources of variation are equivalent.

For the Ollama configuration, persona-associated variation is below both wording and sampling variation. The negative contrasts persist with the second encoder, image bootstrap, available-pair support, and leave-one-image-out means. Every secondary metric also gives a negative persona-minus-wording contrast on the primary complete-image set. The difference in direction between configurations is retained when the models are compared on their 25 shared complete images. These results concern the specified role assignments, wordings, and requested seeds; they do not rank general affordance competence.

\section{Discussion}
\subsection{What the correction changes}
The historical records contain output differences alongside a measurable availability artifact. Correcting the artifact raises overlap; it does not show that the remaining reports are accurate, nor that personas have no effect. The appropriate inference depends on the observation being compared. An empty answer belongs in an availability analysis. A change from one visible object to another belongs in a selection analysis. A changed phrase for the same function belongs in a wording analysis. Treating all three as a common amount of ``affordance drift'' loses precisely the distinctions needed to interpret the result.

The Tucker check illustrates a related problem for representation claims. A stable factor can describe regular response patterns in an output pipeline. Resampling images preserves a persona-specific empty-response mechanism, so bootstrap stability cannot by itself establish the factor's functional meaning. Conversely, the corrected analysis need not find a new latent label to be useful. Its result is that the original evidence does not support the published manifold interpretation.

\subsection{What the new controls identify}
The new experiment isolates a narrower comparison than the original seven-prime study. Within a wording version, the visible image, shared question, output length request, and schema are fixed while the persona sentence changes. The observed contrast concerns the effect of that textual role assignment on the report. Comparing it with wording and sampling variation gives the magnitude an empirical reference, rather than judging low overlap against an arbitrary threshold.

The observed balance differs between configurations. Qwen's persona distances modestly exceed its sampling baseline and are close to its wording baseline; Ollama's persona distances are lower than both. Thus the controls do not support a uniform claim that persona-associated variation dominates report variability. Because the new Qwen arm uses different weights, images, and output constraints from the historical collection, this contrast cannot quantify how much of the old result would disappear under a matched re-run of that model.

Greater persona-associated variation would not itself indicate better perception or greater affordance competence. A model can follow a role stereotype, select a different visible object, describe the same object with a different name, or alter the proposed use. Names-only and uses-only metrics help describe report differences but cannot adjudicate their truth. A small persona-minus-wording contrast is also compatible with substantial absolute report variation; it does not establish global persona invariance.

\subsection{Limits and the next evidential step}
The 48 new images, four personas, two wordings, two requested seeds, one temperature, and two quantized model configurations provide a bounded study. They do not establish a general ranking of VLMs, describe all persona formulations, or predict behavior under different decoding temperatures. The older 479-image data cover more scenes but retain their task-confounded prompts and incomplete model provenance. These datasets answer different questions and are not pooled as a single larger experiment.

The new images were absent from the historical pilot, not necessarily from model training. COCO is a widely used research resource, and the MiniLM model card lists COCO captions among its text training sources \citep{minilmcard}. The experiment measures within-image differences in reports rather than unseen-image generalization. It supplies no claim of uncontaminated visual generalization. A descriptive input-length audit added after collection finds 26 of 2850 nonempty historical Qwen reports and 12 of 3307 nonempty historical Ollama reports above MiniLM's 256-token input limit. Among 592 primary-analysis reports for Qwen3.5-9B, 0 exceed the MiniLM limit and 0 exceed MPNet's 384-token limit. Among 512 primary-analysis reports for Ollama llava:13b, 0 exceed the MiniLM limit and 0 exceed MPNet's 384-token limit. These counts describe the actual encoder inputs; they do not change response inclusion or the primary metric.

There is no independently labeled affordance inventory and no blinded human judgment of object visibility or use feasibility. A useful next study would hold object identity fixed using externally verified regions, ask matched questions about the same object, and evaluate uses against an explicit capability model or human annotation rubric. That would separate selection of different objects from interpretation of a fixed object. Embodied task success would require an additional action-execution study. Neither task is silently replaced here by automatic similarity scoring.

\section{Conclusion}
Context-conditioned visual reports require an analysis that separates response availability from content variation. In the historical pilot, empty answers lowered lexical overlap and created a concentrated input pattern underlying a named latent factor. Reconstructing the pipeline changes the overlap estimates and removes support for the manifold interpretation. In the new matched-question experiment, persona-associated variation is modestly above sampling variation and close to wording variation for Qwen3.5-9B, but below both comparators for the Ollama configuration. The controlled results therefore qualify the earlier generalization about persona effects while retaining a measurable, configuration-dependent result about reports. They establish no percentage of context-dependent meaning, temporal order of visual computation, or improvement in robotic action.

\section*{Data, code, and author responsibility}
The accompanying reproduction archive contains raw model outputs, input and model hashes, the frozen prompt schedule, analysis scripts, generated tables and figures, and verification records. COCO images are identified by filenames and hashes rather than relicensed as part of this archive. The correction record for this revision appears in Appendix~\ref{app:corrections}. Earlier versions remain available under arXiv:2603.04419 and Zenodo concept DOI \href{https://doi.org/10.5281/zenodo.18091089}{10.5281/zenodo.18091089}.

OpenAI Codex assisted with the September reanalysis, implementation, source checks, and manuscript revision. The earlier project also used AI assistance as disclosed in its prior version. The author is responsible for verifying and approving the claims and final submission. No human-participant data were collected for the new experiment, and no new human annotation study is claimed.

\bibliographystyle{plainnat}
\bibliography{references}

\appendix
\section{Full matched-question prompts and execution settings}
\label{app:prompts}
\subsection*{Wording 1}
\noindent\textbf{Neutral}. You are an adult observer. Identify exactly three clearly visible objects in this image, in order of relevance to your perspective. For each, give its name and one physically plausible use supported by its visible properties. Do not invent objects to suit your role. If a role-specific use is unavailable, give an ordinary use of a visible object. Return only JSON with this schema: \{"objects":[\{"name":"object name","use":"one short phrase"\}]\}. Do not include explanations or additional fields.

\noindent\textbf{Chef}. You are a professional chef. Identify exactly three clearly visible objects in this image, in order of relevance to your perspective. For each, give its name and one physically plausible use supported by its visible properties. Do not invent objects to suit your role. If a role-specific use is unavailable, give an ordinary use of a visible object. Return only JSON with this schema: \{"objects":[\{"name":"object name","use":"one short phrase"\}]\}. Do not include explanations or additional fields.

\noindent\textbf{Security}. You are a security professional. Identify exactly three clearly visible objects in this image, in order of relevance to your perspective. For each, give its name and one physically plausible use supported by its visible properties. Do not invent objects to suit your role. If a role-specific use is unavailable, give an ordinary use of a visible object. Return only JSON with this schema: \{"objects":[\{"name":"object name","use":"one short phrase"\}]\}. Do not include explanations or additional fields.

\noindent\textbf{Mobility}. You are a wheelchair user. Identify exactly three clearly visible objects in this image, in order of relevance to your perspective. For each, give its name and one physically plausible use supported by its visible properties. Do not invent objects to suit your role. If a role-specific use is unavailable, give an ordinary use of a visible object. Return only JSON with this schema: \{"objects":[\{"name":"object name","use":"one short phrase"\}]\}. Do not include explanations or additional fields.

\subsection*{Wording 2}
\noindent\textbf{Neutral}. Take the perspective of an adult observer. Name exactly three objects that are plainly visible in the image, ordering them by relevance from your perspective. For each object, state one physically possible use justified by what is visible. Do not add unseen objects because of your role. Use an everyday function of a visible object when no role-specific function applies. Return only JSON with this schema: \{"objects":[\{"name":"object name","use":"one short phrase"\}]\}. Do not include explanations or additional fields.

\noindent\textbf{Chef}. Take the perspective of a professional chef. Name exactly three objects that are plainly visible in the image, ordering them by relevance from your perspective. For each object, state one physically possible use justified by what is visible. Do not add unseen objects because of your role. Use an everyday function of a visible object when no role-specific function applies. Return only JSON with this schema: \{"objects":[\{"name":"object name","use":"one short phrase"\}]\}. Do not include explanations or additional fields.

\noindent\textbf{Security}. Take the perspective of a security professional. Name exactly three objects that are plainly visible in the image, ordering them by relevance from your perspective. For each object, state one physically possible use justified by what is visible. Do not add unseen objects because of your role. Use an everyday function of a visible object when no role-specific function applies. Return only JSON with this schema: \{"objects":[\{"name":"object name","use":"one short phrase"\}]\}. Do not include explanations or additional fields.

\noindent\textbf{Mobility}. Take the perspective of a wheelchair user. Name exactly three objects that are plainly visible in the image, ordering them by relevance from your perspective. For each object, state one physically possible use justified by what is visible. Do not add unseen objects because of your role. Use an everyday function of a visible object when no role-specific function applies. Return only JSON with this schema: \{"objects":[\{"name":"object name","use":"one short phrase"\}]\}. Do not include explanations or additional fields.

\FloatBarrier
\section{Response outcomes by persona}
\begin{table}[htbp]\centering\small
\caption{Response outcomes by persona in the new experiment.}
\label{tab:response-summary}
\begin{tabular}{llrrrrr}
\toprule
Model & Persona & Compliant & Empty & Invalid & Other & Limit stops \\
\midrule
Qwen3.5-9B & Neutral & 191 & 0 & 1 & 0 & 0 \\
 & Chef & 188 & 4 & 0 & 0 & 0 \\
 & Security & 191 & 0 & 1 & 0 & 0 \\
 & Mobility & 187 & 4 & 1 & 0 & 0 \\
Ollama llava:13b & Neutral & 181 & 0 & 3 & 8 & 0 \\
 & Chef & 189 & 0 & 0 & 3 & 0 \\
 & Security & 188 & 0 & 3 & 1 & 0 \\
 & Mobility & 186 & 0 & 1 & 5 & 0 \\
\bottomrule\end{tabular}
\par\smallskip\begin{minipage}{.98\linewidth}\footnotesize Each row has 192 scheduled requests. Invalid combines parse and schema errors. Other includes transport, missing-field, cardinality, or duplicate-name outcomes, whose exact categories are in the response CSV. Limit stops are backend finish reasons and may overlap the outcome columns; they are not automatically excluded when the required content is valid.\end{minipage}
\end{table}

\clearpage
\section{Correction record relative to the earlier version}
\label{app:corrections}
\begin{longtable}{p{.25\linewidth}p{.66\linewidth}}
\toprule Earlier interpretation & Revised treatment \\
\midrule\endhead
Empty objects excluded & The historical implementation retained 363 empty lists. Availability and conditional content are now reported separately; generated correction tables identify both estimands. \\[.5em]
More than 90\% of scene description is context-dependent & Report the lexical similarity and its complement as lexical scores. They are neither the fraction of scene content nor the fraction of affordances altered. \\[.5em]
58.5\% semantic drift and a 40\% invariant semantic core & Withdraw both percentage-of-meaning interpretations. Regenerate cosine scores and report their scale and input-processing limits. \\[.5em]
Culinary Manifold / Access Axis & Reproduce the historical factor numerically, report missing-cell and complete-nonempty sensitivities, and withdraw functional-manifold labels. \\[.5em]
Orthogonality as functional evidence & Orthogonality is imposed by HOOI; uncentered reconstruction-energy capture is distinguished from explained centered variance. \\[.5em]
LLaVA-1.5-13B replication and architectural explanation & Use the historical recorded Ollama tag with unresolved release identity. Pin the current model for the new experiment; make no architectural attribution. \\[.5em]
An arbitrary $J=0.5$ significance test & Replace with image-level persona-minus-wording and persona-minus-sampling contrasts. Retrospective similarity tables are descriptive. \\[.5em]
Visual Genome human baseline & Remove from the evidential comparison: extraction was not reproducible from the released code, task constraints were unmatched, and summary artifacts had inconsistent counting bases. No matched human baseline is claimed. \\[.5em]
Historical stochastic baseline as identification of persona alone & Preserve the archived files for provenance, but replace its inferential role with the new shared-task controls. The old prompts still changed the question and the files did not pin model identity. \\[.5em]
Semantic-first ordering and JIT-ontology robotics claims & Remove the proposed pipeline as an empirical finding. Output behavior does not identify internal order, and no robotic action was tested. \\
\bottomrule
\end{longtable}

\section{Reproduction and audit boundaries}
The analysis scripts reject duplicate response keys and incomplete or mismatched schedules. Frozen input hashes are verified; malformed JSON is not repaired. A regression suite checks empty-response handling, pair definitions, and image weighting. The historical word-score universe matches the archived 9,244-row CSV, and regenerated cosine discrepancies are recorded. Tables, figures, and numerical macros are generated from analysis outputs.

The package pins dependencies and model identities for replaying stored responses; it does not guarantee bitwise regeneration of VLM outputs on other hardware. Bundled bibliography and figures support offline LaTeX compilation. Separate verification records document source-archive compilation, numerical checks, and analysis replay.
\end{document}